\documentclass[10pt,twocolumn,letterpaper]{article}

\usepackage{titling}
\usepackage{cvpr}
\usepackage{times}
\usepackage{epsfig}
\usepackage{graphicx}
\usepackage{booktabs}
\usepackage{url}            
\usepackage{booktabs}       
\usepackage{amsfonts}       
\usepackage{nicefrac}       
\usepackage{microtype}      
\usepackage{amsmath}
\usepackage{amssymb}
\usepackage{mathtools}
\usepackage{color}

\usepackage{lib}
\usepackage{mathtools}
\usepackage{multirow}
\usepackage{dsfont}
\usepackage{balance}
\usepackage{subfigure}

\usepackage[font=small,labelfont=bf]{caption}

\usepackage[pagebackref=true,breaklinks=true,letterpaper=true,colorlinks,bookmarks=false]{hyperref}

\cvprfinalcopy 

\newcommand\blfootnote[1]{%
  \begingroup
  \renewcommand\thefootnote{}\footnote{#1}%
  \addtocounter{footnote}{-1}%
  \endgroup
}

\def\cvprPaperID{000} 

\begin{document}

\title{Multi-view Consistency as Supervisory Signal \\ for Learning Shape and Pose Prediction}

\author{Shubham Tulsiani, Alexei A. Efros, Jitendra Malik\\
University of California, Berkeley \\
{\tt\small \{shubhtuls, efros, malik\}@eecs.berkeley.edu}
}

\maketitle

\begin{abstract}
We present a framework for learning single-view shape and pose prediction without using direct supervision for either. Our approach allows leveraging multi-view observations from unknown poses as supervisory signal during training. Our proposed training setup enforces geometric consistency between the independently predicted shape and pose from two views of the same instance. We consequently learn to predict shape in an emergent canonical (view-agnostic) frame along with a corresponding pose predictor.  We show empirical and qualitative results using the ShapeNet dataset and observe encouragingly competitive performance to previous techniques which rely on stronger forms of supervision. We also demonstrate the applicability of our framework in a realistic setting which is beyond the scope of existing techniques: using a training dataset comprised of online product images where the underlying shape and pose are unknown.
\end{abstract}

\blfootnote{\noindent Project website with code: \footnotesize{\url{https://shubhtuls.github.io/mvcSnP/}}}

\vspace{-2mm}
\section{Introduction}
\vspace{-2mm}
Consider the flat, two-dimensional image of a chair in \figref{teaser}(a). A human observer cannot help but perceive its 3D structure. Even though we may have never seen this particular chair before, we can readily infer, from this single image, its likely 3D shape and orientation. 
To make this inference, we must rely on our knowledge about the 3D structure of other, previously seen chairs. 
But how did we acquire this knowledge? And can we build computational systems that learn about 3D in a similar manner?

Humans are moving organisms: our ecological supervision~\cite{gibson1979ecological} comprises of observing the world and the objects in it from different perspectives, and these multiple views inform us of the underlying geometry. This insight has been successfully leveraged by a long line of geometry-based reconstruction techniques. However these structure from motion  or multi-view stereo methods work for specific instances and  do not, unlike humans, generalize to predict the 3D shape of a novel instance given a single view. Some recent learning-based methods~\cite{choy20163d,Girdhar16b} have attempted to address single-view 3D inference task, but this ability has come at a cost. These approaches rely on full 3D supervision and require known 3D shape for each training image. Not only is this form of supervision ecologically implausible, it is also practically tedious to acquire and difficult to scale. Instead, as depicted in \figref{teaser}(b), our goal is to learn 3D prediction using the more naturally plausible multi-view supervision.

\begin{figure}[t]
\centering
\includegraphics[width=0.48\textwidth]{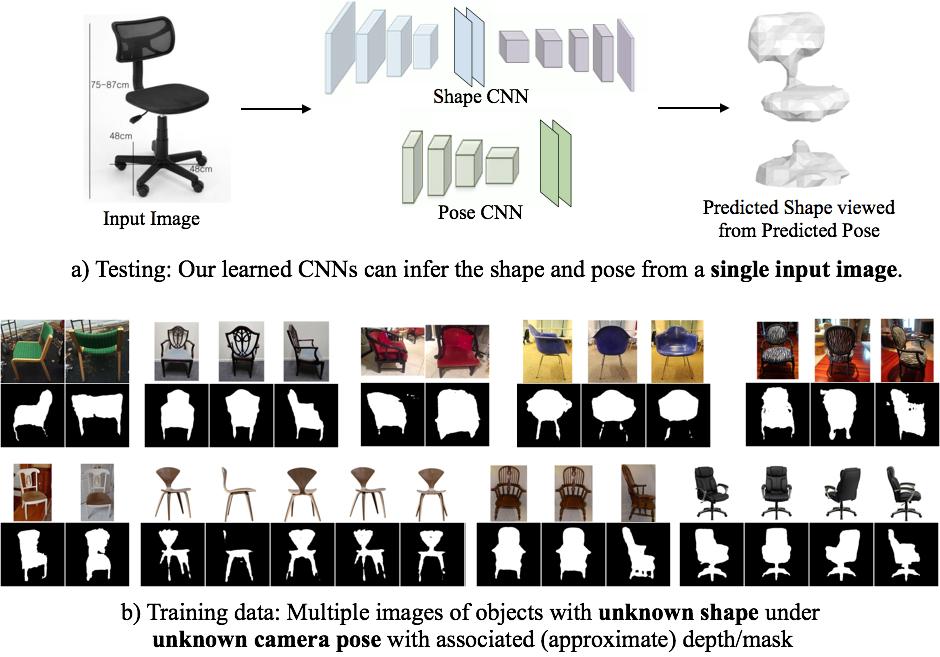}
\vspace{-4mm}
\caption{\small We learn to predict the shape and pose of an object from a single input view. Our framework  can leverage training data of the form of multi-view observations of objects, and learn shape and pose prediction despite the lack of any direct supervision.}
\vspace{-2mm}
\figlabel{teaser}
\end{figure}

\begin{figure*}[t!]
\centering
\includegraphics[width=0.92\textwidth]{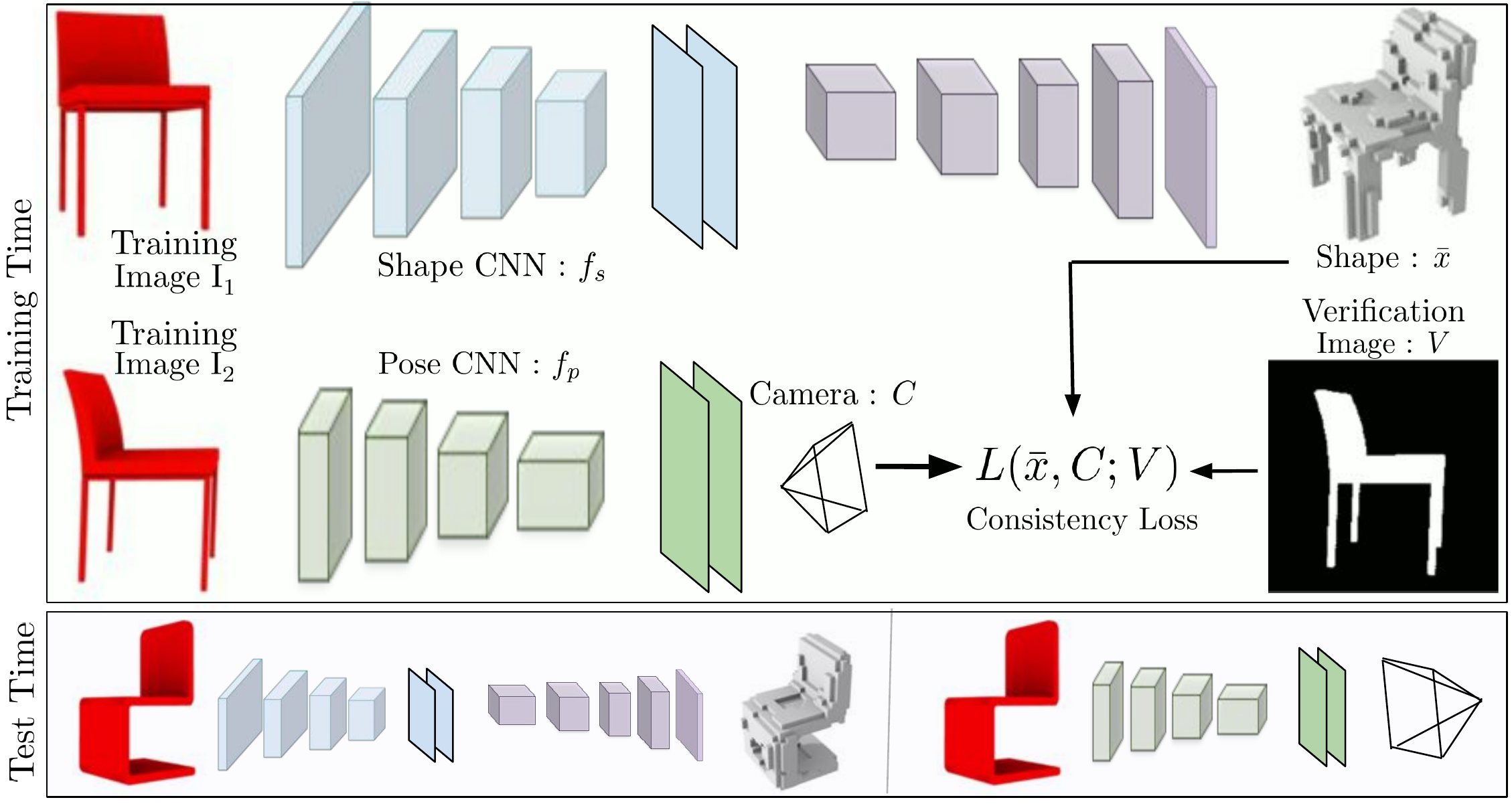}
\vspace{-2mm}
\caption{\small Overview of our approach. During training, we use paired views of the same instance along with a depth/mask verification image from the second view. We predict shape from the first image and pose from the second, and enforce consistency between the shape, pose and the verification image. At test time, our learned models are used to infer the shape and pose from a single RGB input image.}
\vspace{-2mm}
\figlabel{overview}
\end{figure*}

The broader goal of learning from data without explicit supervision is the focus of
of considerable attention in the deep learning literature. Mechanisms that have been proposed include the use of information bottlenecks or proxy tasks such as prediction that encourage learning about the temporal or spatial structure. Similarly, in this paper, we rely on enforcing a geometric bottleneck for the task of explaining novel views and leverage the principle of multi-view consistency: a common geometry, observed from different perspectives can consistently explain multiple views of an instance. While some recent approaches~\cite{rezende2016unsupervised,drcTulsiani17,yan2016perspective} have utilized these principles to learn 3D shape prediction, they all crucially rely on object pose supervision during training. Our proposed framework allows us to go a step further, and learn single-view shape and pose prediction using multi-view observations from \emph{unknown} poses. Therefore, unlike previous methods which require either shape or pose supervision, we relax the requirement for \emph{both} these forms of supervision.

Our approach, as summarized in \figref{overview}, learns shape and pose prediction by enforcing consistency between the predictions and available (novel view) observations. Concretely, given one image of an object instance, we predict a corresponding shape. In parallel, given a \emph{different} image of the same instance, we independently predict a corresponding pose. Then, we enforce that the predicted shape (using the former image) should be `consistent' with a depth/mask observation for the latter image when viewed from the predicted pose. As we discuss in \secref{learning}, and demonstrate qualitatively and quantitatively demonstrate in \secref{experiments}, this allows us to learn single-view shape and pose prediction despite not having direct supervision for either.

\vspace{-2mm}
\section{Related Work}
\vspace{-2mm}
\seclabel{related}
\noindent \textbf{Structure from Motion and Multi-view Instance Reconstruction.}
Structure from motion (SfM)~\cite{ullman1979interpretation} based methods (e.g.~\cite{brown2005unsupervised,snavely2006photo}) aim to recover the geometry, typically as sparse 3D point clouds, and the camera pose for each image. It was also shown that volumetric representations can be inferred by fusing multiple range images~\cite{curless1996volumetric} or foreground masksl~\cite{broadhurst2001probabilistic,laurentini1994visual,matusik2000image}. More closely related to our formulation, ray-potential based optimization methods~\cite{de1999roxels,liu2010ray} can be used to infer discrete or probabilistic~\cite{ulusoy2015towards} volumetric representations from multiple color images. This class of optimization techniques can be further extended to incorporate additional signals \eg depth or semantics~\cite{kundu2014joint,savinov2016semantic,savinov2015discrete}. The goal of all these multi-view instance reconstruction methods is to infer the 3D structure of a specific scene/object given a large number of views of the {\em same instance}. Our method can be thought of as trying to minimize similar cost functions during training, but at test time, we can infer the pose and shape from a \emph{single RGB image} -- something that these classical techniques cannot do.

\vspace{1mm}
\noindent \textbf{Generative 3D Modeling without 3D Supervision.}
Blanz and Vetter~\cite{blanz1999morphable}, using 3D supervision, captured the shapes of faces using a deformable model. Cashman and Fitzgibbon~\cite{cashman2013shape} subsequently demonstrated that similar generative models could be learned using only image based annotations. Kar \etal~\cite{shapesKarTCM15} extended these ideas to more general categories and automated test-time inference using off-the shelf recognition systems. However, these models are restricted to only capture deformations around a mean shape(s), thus limiting their expressiveness. Recently, Gadhela \etal~\cite{gadelha20163d} presented a more expressive generative model for shapes learned using a collection of silhouette images but did not examine applications for inference conditioned on image evidence. Eslami \etal~\cite{eslami2016attend} also learned a generative model with a corresponding inference module using only RGB images but only demonstrated 3D inference in scenarios where object shapes were known a priori. While the recent successes indicate that multi-view (or even single-view) observations can allow learning expressive generative models, their applications for single-view reconstruction have not been demonstrated conclusively. We instead propose to discriminatively train single-view shape and pose estimation systems using similar multi-view observations.

\vspace{1mm}
\noindent \textbf{Multi-view Supervision for Single-view Depth Prediction.}
A recent direction pursued in the area of learning-based single-view depth prediction is to forego the need for direct supervision~\cite{eigen2015predicting} and instead rely on multi-view observations for training~\cite{garg2016unsupervised,godard2016unsupervised,zhou2017unsupervised}.
Garg \etal~\cite{garg2016unsupervised} and Godard \etal~\cite{godard2016unsupervised} leverage stereo images as supervision to learn single image depth prediction. Zhou \etal~\cite{zhou2017unsupervised} further relax the assumption of known relative pose between the multiple views, and learn single-view depth and ego-motion prediction models from monocular videos. Similarly, we leverage multiple views from unknown poses as supervisory signal but we pursue 3D instead of 2.5D predictions.


\vspace{1mm}
\noindent \textbf{Multi-view Supervised Single-view Reconstruction.}
Initial CNN-based methods~\cite{choy20163d, Girdhar16b,marrnet} predicted voxel occupancy representations from a single input image but required full 3D supervision during training. Recent approaches have advocated using alternate forms of supervision. Zhu \etal~\cite{zhu17reproj} showed that systems trained using synthetic shape and pose supervision could be adapted to real data using only image based annotation. Their pre-training, however, crucially relied on direct shape and pose supervision. Towards relaxing the need of any shape supervision, some recent methods demonstrated the feasibility of using multi-view foreground masks~\cite{gwak2017weakly,rezende2016unsupervised,yan2016perspective} or more general forms of observation \eg depth, color, masks, semantics \etc~\cite{drcTulsiani17} as supervisory signal.
Our work adheres to this ideology of using more natural forms of supervision for learning 3D prediction and we take a step further in this direction. The previous multi-view supervised approaches~\cite{gwak2017weakly,rezende2016unsupervised,drcTulsiani17,yan2016perspective} required known camera poses for the multiple views used during training and our work relaxes this requirement.

\vspace{-2mm}
\section{Approach}
\vspace{-2mm}
\seclabel{learning}
We aim to learn shape and pose prediction systems, denoted as $f_s$ and $f_p$ respectively, which can infer the corresponding property for the underlying object from a single image. However, instead of direct supervision, the supervision available is of the form of multi-view observations from unknown poses. We first formally define our problem setup by describing the representations inferred and training data leveraged and then discuss our approach.

\vspace{1mm}
\noindent \textbf{Training Data.}
We require a sparse set of multi-view observations for multiple instances of the same object category. Formally, denoting by $\mathcal{N}(i)$ the set of natural numbers up to $i$, we assume a dataset of the form $\{ \{(I^i_v, V^i_v)~|~v \in \mathcal{N}(N_i) \}~|~i \in \mathcal{N}(N) \}$. This corresponds to  $N$ object instances, with $N_i$ views available for the $i^{th}$ instance. Associated with each image $I_i^v$, there is also a depth/mask image $V_i^v$ that is used for consistency verification during training. Note that there is no direct pose or shape supervision used -- only multi-view observations with identity supervision.

\noindent \textbf{Shape and Pose Parametrization.}
The (predicted) shape representation $\bar{x}$ is parametrized as occupancy probabilities of cells in a 3D grid. The pose of the object, parametrized as a translation $t$ and rotation $R$, corresponds to the camera extrinsic matrix. While we assume known camera intrinsics for our experiments, our framework can also be extended to predict these.

\vspace{-1mm}
\subsection{Geometric Consistency as Supervision}
\vspace{-1mm}
Multiple images of the same instance are simply  renderings of a common geometry from diverse viewpoints. Therefore, to correctly `explain' multiple observations of an instance, we need  the correct geometry (shape) of the instance and the corresponding viewpoints (pose) for each image. Our approach, which is depicted in \figref{overview}, builds on this insight and proposes to predict \emph{both}, shape and pose s.t. the available multi-view observations can be explained.

Concretely, during training, we use one image of an instance to predict the instance shape. In parallel, we use a \emph{different} image of the same instance to predict pose. Then, we enforce that the predicted shape, when viewed according to the predicted pose, should be consistent with a depth/mask image from the latter view. We therefore use the notion of \emph{consistency} as a form of meta-supervision \ie while the ground-truth shape and pose are unknown, we know that they should be consistent with the available verification image. After the training stage, our learned models can infer shape and pose from a single view of a novel instance. 

A crucial aspect of the designed training setup is that the shape and pose estimates are \emph{independently} obtained from \emph{different} images of the same instance. This enforces that the optimal solution corresponds to predicting the correct shape and pose. Another interesting property is that the shape is predicted in an emergent canonical, view-independent frame, and the predicted pose is with respect to this frame.

\vspace{1mm}
\noindent \textbf{Correctness of Optimal Shape and Pose.}
We consider \figref{overview} and first examine the shape prediction CNN $f_s$. It predicts a shape $f_s(I_1)$ given some input image. This shape is verified against $V$ from a different view which is unknown to $f_s$. The optimal predicted shape should therefore be consistent with \emph{all} possible novel views of this instance, and therefore correspond to the true shape (upto some inherent ambiguities \eg concavities in case of mask supervision). Similarly, the pose prediction CNN $f_p$ is required to infer a viewpoint under which the predicted geometry can explain the verification image $V$. As  $V$ is chosen to be from the same viewpoint as the image $I_2$, the pose CNN should  predict the correct viewpoint corresponding to its input image ($I_2$).

\vspace{1mm}
\noindent \textbf{Emergent Canonical Frame.}
Under our proposed setup, the predicted pose $f_p(I_2)$ is agnostic to the image $I_1$. However, to explain the verification image $V$, the pose CNN is required to predict a pose w.r.t the inferred shape $f_s(I_1)$. So how can $f_p$ infer pose w.r.t $f_s(I_1)$ when it does not even have access to $I_1$? The resolution to this is that the shape prediction CNN $f_s$ automatically learns to predict shape in some (arbitrary) view-agnostic canonical frame (e.g. `front' of chairs may always face towards the X axis), and the pose CNN $f_p$ learns to predict pose w.r.t this frame. Therefore, even though it is not explicitly enforced, our approach of independently inferring shape and pose makes the learnt CNNs automatically adhere to some emergent canonical frame.

Towards implementing our framework, we require a consistency loss $L(\bar{x}, C; V)$ which measures whether the (predicted) shape $\bar{x}$ and camera pose $C$ can geometrically explain a depth/mask image $V$. We present a formulation for this loss in \secref{pdrc} and then describe the training process in \secref{training}. We finally describe some modifications required to make the training more robust.

\vspace{-1mm}
\subsection{Pose-differentiable Consistency Loss}
\vspace{-1mm}
\seclabel{pdrc}
We formulate a view consistency loss $L(\bar{x},C; V)$ that measures the inconsistency between a shape $\bar{x}$ viewed according to camera $C$ and a depth/mask image $V$. Our formulation builds upon previously proposed differentiable ray consistency formulation~\cite{drcTulsiani17}. However, unlike the previous formulation, our proposed view consistency loss is differentiable w.r.t pose (a crucial requirement for usage in our learning framework). Here, we very briefly recall the previous formulation and mainly highlight our proposed extension. A more detailed and complete formulation of the view consistency loss can be found in the appendix.

\vspace{1mm}
\noindent \textbf{Differentiable Ray Consistency~\cite{drcTulsiani17}.} The view consistency loss formulated by Tulsiani \etal~\cite{drcTulsiani17} could be decomposed into per-pixel (or ray) based loss terms where $L_p(\bar{x},C; v_p)$ denotes the consistency of the shape and camera with the observation $v_p$ at pixel $p$. The per-pixel loss is defined as the \emph{expected event cost}:
\vspace{-2mm}
\begin{gather}
L_p(\bar{x},C; v_p)~~=~~\sum_{i=1}^{N} q_p(i) \psi_p(i)
\vspace{-2mm}
\end{gather}
Here, $\psi_p(i)$ denotes the cost for each event, determined by $v_p$, and $q_p(i)$ indicates the \emph{event probability} \ie the likelihood of the ray stopping at the $i^{th}$ voxel in its path. The event probability, $q_p(i)$ is in turn instantiated using the probabilities $\{x_p^i\}$ - where $x_p^i$ denotes the occupancy probability of the $i^{th}$ voxel in the ray's path. See appendix for details.

\vspace{1mm}
\noindent \textbf{Sampling Occupancies along a Ray.}
The loss function as defined above is differentiable w.r.t shape $\bar{x}$, but not the camera parameters. This is because the quantity $\{x_p^i\}$  is not a differentiable function of the camera (since the ordering of voxels on a ray's path is a discrete function). Our insight is that instead of looking up \emph{voxels} on the ray's path, we can consider \emph{samples} along its path. Thus, our formulation is similar to that proposed by Tulsiani \etal~\cite{drcTulsiani17}, with the difference that the variable $\{x_p^i\}$ is redefined to correspond to the occupancy at the $i^{th}$ point sample along the ray.

Concretely, we sample points at a fixed set of $N = 80$ depth values $\{d_i | 1 \leq i \leq N\}$ along each ray. To determine $x^p_i$, we look at the 3D coordinate of the corresponding point (determined using camera parameters), and trilinearly sample the shape $\bar{x}$ to determine the occupancy at this point.
\vspace{-2mm}
\begin{gather}
l_i \equiv (\frac{u-u_0}{f_u}d_i, \frac{v-v_0}{f_v}d_i,d_i) \\
x^p_i = \mathcal{T}( \bar{x} , R \times (l_i + t)~)
\end{gather}
As the trilinear sampling function $\mathcal{T}$ is differentiable w.r.t its arguments, the sampled occupancy $x^p_i$ is differentiable w.r.t the shape $\bar{x}$ and the camera $C$. We note that Yan \etal~\cite{yan2016perspective} also used a similar sampling trick but their formulation is restricted to specifically using mask verification images and is additionally not leveraged for learning about pose.

\vspace{-1mm}
\subsection{Learning}
\seclabel{training}
\noindent \textbf{Training Objective.}
To train the shape and pose predictors, we leverage the view consistency loss previously defined (\secref{pdrc}) and train $f_s, f_p$ jointly to minimize $L_{data} = \sum\limits_{i=1}^{N}~\sum\limits_{u=1}^{N_i}~\sum\limits_{v=1}^{N_i}~ L(f_s(I^i_u), f_p(I^i_v); V^i_v)$. Therefore, the shape predicted using every image $f_s(I^i_u)$ should be consistent with \emph{all} available verification images of the same instance ($\{V^i_v\}$) when viewed from the corresponding (predicted) poses ($\{f_p(I^i_v)\}$). As detailed earlier, the independent prediction of shape and pose from different images ensures that the CNNs learn to infer the correct shape and pose under some emergent canonical frame.

\vspace{1mm}
\noindent \textbf{Architecture and Optimization Details.} We use a mini-batch size of 8 images $I_u^i$ for which shape is predicted. For each of these images, we randomly sample at least 2, and upto 3 if available, out of $N_i$, views $I_v^i$ of the same instance \ie the mini-batch size for the pose prediction CNN is between 16 and 24. We use extremely simple CNN architectures (depicted in \figref{overview}) corresponding to $f_s$ and $f_p$. Note that both these CNNs are initialized randomly (without any pre-training) and trained using ADAM~\cite{adam}.

\emph{Shape Prediction.} Our shape prediction CNN has an encoder-decoder structure similar to the one used by Tulsiani \etal~\cite{drcTulsiani17}. The input to the CNN is an RGB image of size $64 \times 64$ and the outputs are corresponding voxel occupancy probabilities for a $32 \times 32 \times 32$ grid.

\emph{Pose Prediction.} Our pose prediction CNN $f_p$ has a similar encoder to $f_s$, but outputs the predicted pose via fully connected layers. The rotation aspect of the pose is parametrized using two euler angles (azimuth, elevation) and the predicted translation $\in \mathbb{R}^3$. However, for some analysis experiments, we also assume that the object is at a known location w.r.t the camera and only predict the camera rotation. While in this work we assume known intrinsic parameters, the pose prediction CNN could in principle be extended to infer these.

\begin{figure*}
\centering
\includegraphics[width=0.95\textwidth]{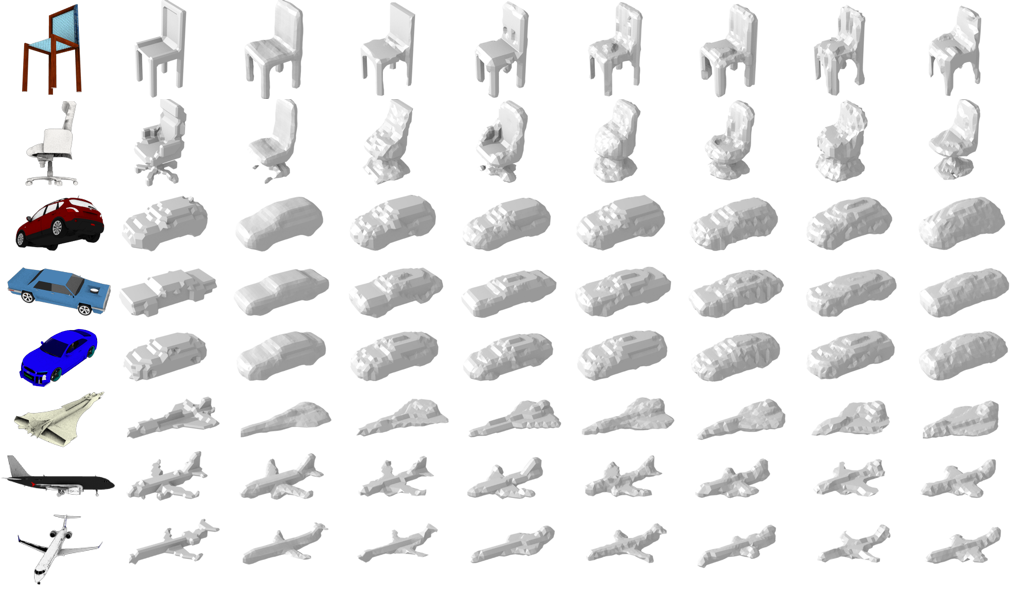}
\vspace{-2mm}
\caption{\small Shape predictions on the validation set using a single RGB input image. We visualize the voxel occupancies by rendering the corresponding mesh (obtained via marching cubes) from a canonical pose. Left to Right: a) Input Image b) Ground-truth c) 3D Supervised Prediction d,e) Multi-view \& Pose Supervision (Mask, Depth) f,g) Mult-view w/o Rotation Supervision (Mask, Depth), and h,i) Mult-view w/o Rotation  and Translation Supervision (Mask, Depth)}
\vspace{-2mm}
\figlabel{shapePred}
\end{figure*}

\vspace{-1mm}
\subsection{Overcoming Local Minima}
\vspace{-1mm}
We observed that our training is susceptible to local minima, in particular for the pose prediction CNN $f_p$. This is not too surprising since we have to learn both shape and pose from scratch, and erroneous estimates for one could confound the learning for the other, particularly in the in the initial stages  We observe that the $f_p$ learns to predict only a small range of poses and \eg instead of predicting back-facing chairs, it confuses them with front-facing chairs. To avoid such local minima, we introduce two changes to the setup previously described.

\vspace{1mm}
\noindent \textbf{Incorporating a Pose Prior.}
We encourage the distribution of the predicted poses to be similar to a prior distribution (uniform azimuth $\in [0,360)$, elevation $\in [-20,40)$ degrees). We do so by adding an adversarial loss for the predictions of $f_p$ where the `real' samples are drawn from the prior distribution and `generated' samples are those predicted by $f_p$. We empirically show that our training is robust to the exact prior and that it can be different from the true distribution.

\vspace{1mm}
\noindent \textbf{Allowing Diverse Predictions.}
While the adversarial loss encourages diverse predictions, we also need some architectural changes  to easily capture these. Instead of directly regressing to a single pose estimate in the last layer, we predict $N_p = 8$ estimates and additionally predict a probability distribution over these. We then sample a pose according to the predicted distribution. We use  Reinforce~\cite{williams1992simple} to obtain gradients for the probability predictions.

\vspace{-2mm}
\section{Experiments}
\vspace{-2mm}
\seclabel{experiments}
We consider two different scenarios where we can learn single-view shape and pose prediction using multi-view observations from unknown poses. We first examine the ShapeNet dataset where we can synthetically generate images and compare our approach against previous techniques which rely on stronger forms of supervision. We then consider a realistic setting where the existing approaches, all of which require either shape or pose supervision, cannot be applied due to lack of any such annotation. Unlike these existing methods, we show that our approach can learn using an online product dataset where multiple images on objects are collected from product websites \eg eBay. 

\subsection{Empirical Analysis using ShapeNet}
\subsubsection{Experimental Setup}
\vspace{-1mm}

\renewcommand{\arraystretch}{1.2}
\setlength{\tabcolsep}{4pt}

\begin{table*}[t]
\centering
\footnotesize
\hfill %
\begin{subtable}{}
    \begin{tabular}{l c cc cc cc}
    \toprule
    Training & \multicolumn{1}{c}{3D} & \multicolumn{2}{c}{{Multi-view}} &  \multicolumn{2}{c}{{Multi-view}} &  \multicolumn{2}{c}{{Multi-view w/o}} \\
    Data & & \multicolumn{2}{c}{{\& GT Pose}} &  \multicolumn{2}{c}{{ w/o Rot}} &  \multicolumn{2}{c}{{Rot \& Trans}}
    \\
     \cmidrule(lr){3-4}
     \cmidrule(lr){5-6}
     \cmidrule(lr){7-8}
     {class} &  & {Mask} & {Depth} & {Mask} & {Depth} & {Mask} & {Depth} \tabularnewline
    \midrule 
    {aero} & 0.57 & 0.55 & 0.43 & 0.52 & 0.44 & 0.38 & 0.37  \tabularnewline
    {car} & 0.79 & 0.75 & 0.69 & 0.74 & 0.71 & 0.48 & 0.68 \tabularnewline
    {chair} & 0.49 & 0.42 & 0.45 & 0.40 & 0.43 & 0.35 & 0.37 \tabularnewline
    \midrule {mean} & 0.62 & 0.57 & 0.52 & 0.55 & 0.53 & 0.40 & 0.47 \tabularnewline
    \bottomrule
    \end{tabular}
\end{subtable}%
\hfill %
\begin{subtable}{}
    \begin{tabular}{l cc cc cc cc cc}
    \toprule
    Training & \multicolumn{2}{c}{GT} &  \multicolumn{4}{c}{MV w/o Rot}  &  \multicolumn{4}{c}{MV w/o Rot \& Trans}
    \\
    Data & \multicolumn{2}{c}{Pose} &  \multicolumn{2}{c}{Mask}  &  \multicolumn{2}{c}{Depth} &  \multicolumn{2}{c}{Mask}  &  \multicolumn{2}{c}{Depth}
    \\
    \cmidrule(lr){2-3}
     \cmidrule(lr){4-5}
     \cmidrule(lr){6-7}
     \cmidrule(lr){8-9}
     \cmidrule(lr){10-11}
     {class} & Acc & Err & Acc & Err & Acc & Err & Acc & Err & Acc & Err \tabularnewline
    \midrule 
    {aero} & 0.79 & 10.7 & 0.69 & 14.3 & 0.60 & 21.7 & 0.53 & 26.9 & 0.63 & 12.3 \tabularnewline
    {car} & 0.90 & 7.4 & 0.87 & 5.2 & 0.85 & 4.9 & 0.53 & 24.8 & 0.56 & 20.6  \tabularnewline
    {chair} & 0.85 & 11.2 & 0.81 & 7.8 & 0.83 & 8.6 & 0.55 & 24.0 & 0.62 & 19.1 \tabularnewline
    \midrule {mean} & 0.85 & 10.0 & 0.79 & 9.0 & 0.76 & 11.7 & 0.54 & 25.1 & 0.61 & 17.4 \tabularnewline
    \bottomrule
    \end{tabular}
\end{subtable}%
\hfill %
\vspace{2mm}
\caption{Analysis of the performance for single-view shape (Left) and pose (Right) prediction. a) Shape Accuracy: Mean IoU on the test set using various supervision settings. b) Pose Accuracy/Error: Acc$_\frac{\pi}{6}$ and Med-Err across different supervision settings.}
\tablelabel{snetEval}
\vspace{-2mm}
\end{table*}

\begin{figure*}
\centering
\includegraphics[width=1.0\textwidth]{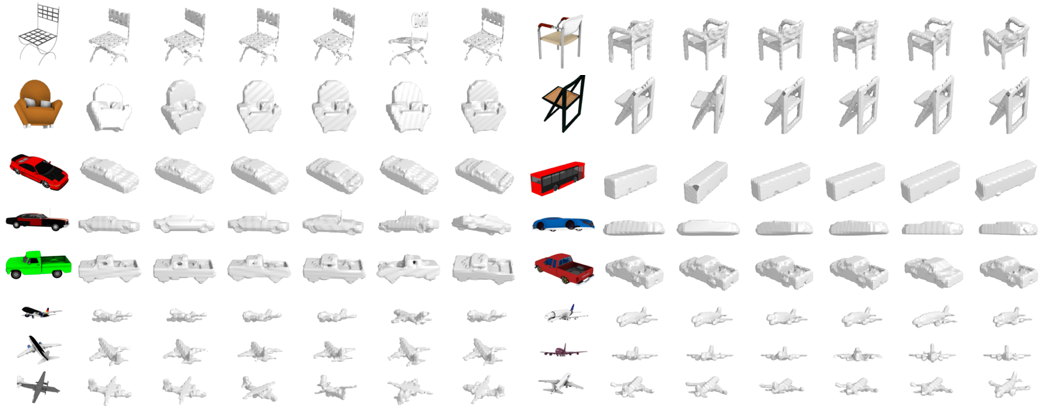}
\caption{\small Rotation predictions on a random subset of the validation images. For visualization, we render the ground-truth voxel occupancies using the corresponding rotation. Left to Right: a) Input Image b) Ground-truth Rotation c) GT Supervised Prediction d,e) Multi-view w/o Rot Supervision (Mask, Depth), and f,g) Multi-view w/o Rot and Trans Supervision (Mask, Depth)}
\vspace{-2mm}
\figlabel{posePred}
\end{figure*}

\noindent \textbf{Dataset.}
We use the ShapeNet dataset~\cite{shapenet2015} to empirically validate our approach. We evaluate on three representative object categories with a large number of models : airplanes, cars, and chairs. We create random train/val/test splits with  $(0.7,0.1,0.2)$ fraction of the models respectively. For each training model, we use $N_i = 5$ images available from different (unknown) views with corresponding depth/mask observations. The images are rendered using blender and correspond to a viewpoint from a randomly chosen azimuth $\in [0,360)$ degrees and elevation $\in [-20,40]$ degrees. We additionally use random lighting variations during rendering.

We also render the training objects under two  settings - a) origin centred, or b) randomly translated around the origin. As the camera is always at a fixed distance away from the origin, the first setting corresponds to training with a known camera translation, but unknown rotation. The second  corresponds to training with both translation and rotation unknown. To have a common test set across various control setting (and compare to~\cite{drcTulsiani17}), we use the origin centered renderings for our validation and test sets. We note that these rendering settings are rather challenging and correspond to significantly more variation than commonly examined by previous multi-view supervised methods which examine settings with fixed translation~\cite{drcTulsiani17}, and sometimes only consider 24~\cite{yan2016perspective} or even 8~\cite{gadelha20163d} possible discrete views.

\begin{figure*}
\centering
\includegraphics[width=1.0\textwidth]{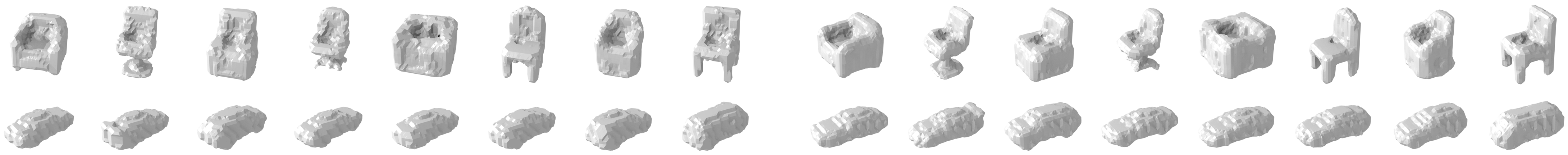}
\vspace{-4mm}
\caption{\small Visualization of 8 random predicted shapes from azimuth $=60^{\circ}$, elevation $=30^{\circ}$. Left: Original predictions from the shape CNN. Right: Shape predictions transformed according to the optimal rotation.}
\vspace{-2mm}
\figlabel{alignment}
\end{figure*}

\begin{figure*}
\centering
\includegraphics[width=1.0\textwidth]{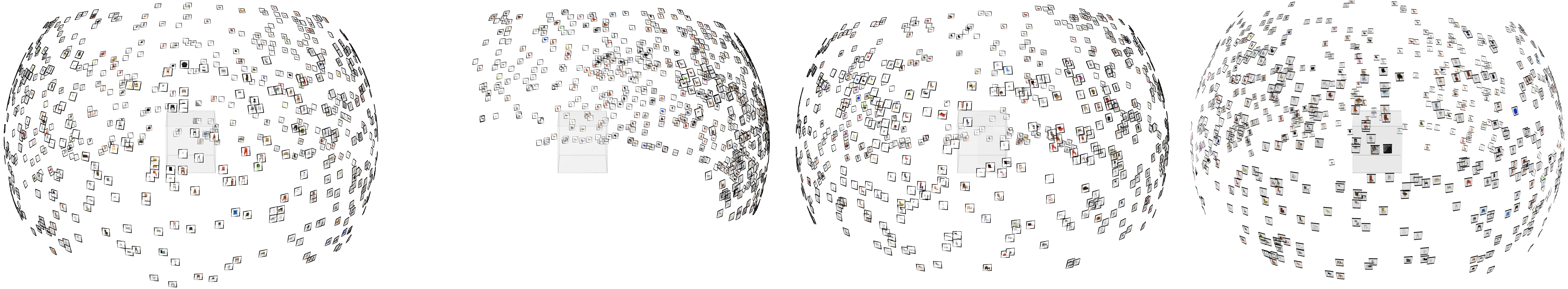}
\vspace{-4mm}
\caption{\small Visualization of the predicted pose distribution under various training settings. Each small image is placed at the (predicted/known) location of the corresponding camera. The reference grid in the centre depicts the space in which shape is predicted. Left to Right : a) Ground-truth poses b) No pose prior c) True pose prior d) Incorrect pose prior, discarded midway through training. See text for details.}
\vspace{-3mm}
\figlabel{posePrior}
\end{figure*}

\vspace{1mm}
\noindent \textbf{Control Settings.} In addition to reporting the performance in the scenario where pose and shape supervision is unavailable, we also examine the settings where stronger supervision \eg shape or pose can be used. These experiments serve to highlight the upper bound performance. In all the experiments, we train a separate model per object category. The various settings studied are : 

\emph{3D Supervision.} To mimic the setup used by 3D supervised approaches~\cite{choy20163d,Girdhar16b}, we assume known ground-truth 3D models for each training image and train the shape CNN using a cross-entropy loss.

\emph{Multi-view with Ground-truth Pose.} In this supervision setting used by previous multi-view supervised approaches, pose (but not shape) supervision is available for the multiple observations. We use our loss function but train the shape prediction CNN $f_s$ using the ground-truth pose instead of predicted poses. We separately train the pose prediction CNN $f_p$ using squared L2 loss in quaternion space (after accounting for antipodal symmetry of quaternions).

\emph{Multi-view without Pose Supervision.} This represents our target setting with the weakest form of supervision available. We train the shape and pose prediction CNNs jointly using our proposed loss. Further, we consider two variants of this setting - one where camera translation is known, one where both camera translation and rotation are unknown.

\vspace{1mm}
\noindent \textbf{Evaluation Metrics.} We report the results using predictions for 2 images per test model. For evaluating the shape prediction CNN, we report the mean intersection over union (IoU) between the ground-truth and predicted shapes. Since different CNNs can be calibrated differently, we search for the optimal threshold (per CNN on the validation set) to binarize the predictions. 
To evaluate the rotation prediction, we measure the angular distance between the predicted and ground-truth rotation (in degrees) and report two metrics : a) Fraction of instances with error less than 30 degrees (Acc$_\frac{\pi}{6}$), and b) Median Angular Error (Med-Err).

\vspace{-1mm}
\subsubsection{Results}
\vspace{-1mm}
\seclabel{mainExp}

\noindent \textbf{Prediction Frame Alignment.} The ShapeNet models are all aligned in a canonical frame where X and Y axes represent lateral and upward directions. The shape and pose prediction CNNs learned using our approach are not constrained to adhere to this frame and in practice, learn to predict shape and pose w.r.t some arbitrary frame.

However, to evaluate these predictions, we compute an optimal rotation to best align the predictions to the canonical ShapeNet frame. We use 8 \emph{random} images per category (the first validation mini-batch) alongwith the ground-truth 3D voxelizations and search for a rotation that maximizes the voxel overlap between the ground-truth and the rotated predicted shapes. We visualize the prediction frame alignment for car and chair CNNs trained using multi-view observations w/o pose via depth verification images in \figref{alignment}. Note that the prediction frames across classes vary arbitrarily. After the alignment process, the predictions for both categories are in the canonical ShapeNet frame.

\begin{figure*}
\centering
\includegraphics[width=1.0\textwidth]{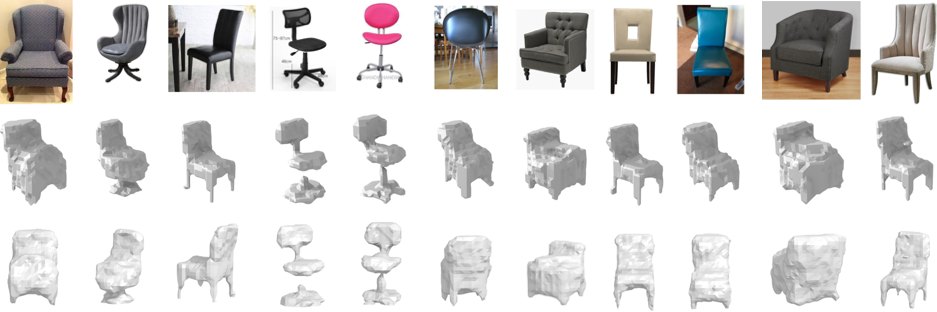}
\vspace{-4mm}
\caption{\small Visualization of predictions using the Stanford Online Product Dataset. (Top) Input image. (Middle) Predicted shape in the emergent canonical pose. (Bottom) Predicted shape rotated according to the predicted pose.}
\vspace{-2mm}
\figlabel{ebay}
\end{figure*}

\vspace{1mm}
\noindent \textbf{Role of a Pose prior.} While the empirical results reported below correspond to using the correct pose prior, we first show that the primary benefit of this prior is that it encourages the CNN to predict diverse poses and avoid local minima, and that even an approximate prior is sufficient. To further support this point, we conducted an experiment where we used an incorrect pose prior (elevation uniform $\in [-40,80]$ instead of $\in [-20,40]$) and removed the prior loss midway through training. We observed that this network also trained successfully, indicating that we do not require the true pose prior, rather only  an approximate one. \figref{posePrior} visualizes the pose distributions inferred under various settings. While using no prior results in a local optima, using the approximate prior (or the correct prior) does not.

\vspace{1mm}
\noindent \textbf{Single-view Shape Prediction.} Our results and the performance under various control settings with stronger supervision is reported in \tableref{snetEval} and visualized in \figref{shapePred}. In general, we observe that the results using our approach are encouragingly close to those obtained using much stronger forms of supervision. This clearly indicates that our approach is able to learn single-view shape prediction despite the lack of either shape or pose information during training. As expected, we also observe that we cannot learn about concavities in chairs via consistency against mask validation images, though we can do so using depth images. e observe a noticeable performance drop in case of mask supervision with unknown translation, as this settings results in scale ambiguities which our evaluation does not account for \eg we learn to predict larger cars, but further away, and this results in a low empirical score.

\vspace{1mm}
\noindent \textbf{Single-view Pose Estimation.} The results of our approach are reported in \tableref{snetEval} and visualized in \figref{posePred}. We observe a similar trend for the task of pose prediction -- that our approach performs comparably to directly supervised learning using ground-truth pose supervision. Interestingly, we often get lower median errors than the supervised setting. We attribute this to the different topologies of the loss functions. The squared L2 loss used in the supervised setting yields small gradients if the pose is almost correct. Our consistency loss however, would want the observation image to perfectly align with the shape via the predicted pose.

\vspace{1mm}
\noindent \textbf{Interpretation.}
The main takeaway from these results is that it is indeed possible to learn shape and pose prediction without direct supervision for either. We empirically and qualitatively observe competitive performances for both these tasks when compared to approaches that leverage stronger forms of supervision. We see that we always learn meaningful shape and pose prediction systems across observation types (mask/depth) and that performance degrades gracefully when using less supervision (known/unknown translation).

\vspace{-1mm}
\subsection{Learning from Online Product Images}
\vspace{-1mm}

\noindent \textbf{Dataset.}
We examined the `chair' object category from the Stanford Online Products Dataset~\cite{song16product} which comprises of automatically downloaded images from eBay.com~\cite{eBay}. Since multiple images (views) of the same product are available, we can leverage our approach to learn from this data. As we also require associated foreground masks for these images, we use an out-of-the-box semantic segmentation system~\cite{chen2017deeplab} to obtain these. However, the obtained segmentation masks are often incorrect. Additionally, many of the product images were not suited for our setting as they only comprised of a zoom-in of a small portion of the instance (\eg chair wheel). We therefore manually selected images of unoccluded/untruncated instances with a reasonably accurate (though still noisy) predicted segmentation. We then used the object instances with atleast 2 valid views for training. This results in a filtered dataset of $N = $ 282 instances with $N_i = $  3.65 views on average per instance.

\vspace{1mm}
\noindent \textbf{Results.}
We can apply our approach to learn from this  dataset comprising of multiple views with associated (approximate) foreground masks. Since the camera intrinsics are unknown, we assume a default intrinsic matrix (see appendix). We then learn to predict the (unknown) translation and rotation via $f_p$ and the (unknown) shape via $f_s$ using the available multi-view supervision. Note that the learned CNNs are trained from scratch, and that we use the same architecture/hyperparameters as in the ShapeNet experiments.

Some results (on images of novel instances) using our learned CNN are visualized in \figref{ebay}.  We see that we can learn to predict meaningful 3D structure and infer the appropriate shape and pose corresponding to the input image. Since only foreground mask supervision is leveraged, we cannot learn to infer the concavities in shapes. We also observe confusion across poses which result in similar foreground masks. However, we feel that this result using training data derived from a challenging real world setting, concretely demonstrates our method's ability to learn despite the lack of direct shape or pose supervision. To the best of our knowledge, this is the first such result and it represents an encouraging step forward.

\vspace{-2mm}
\section{Discussion}
\vspace{-1mm}
We presented a framework that allows learning single-view prediction of 3D structure without direct supervision for shape or pose. While this is an encouraging result that indicates the feasibility of using natural forms of supervision for this task, a number of challenges remain to be addressed. As our supervisory signal, we rely on consistency with validation images of unoccluded objects and it would be useful to deal with unknown occlusions. It would also be interesting to apply similar ideas for learning the 3D structure of general scenes though this might additionally require leveraging alternate 3D representations and allowing for object motion to handle dynamic scenes.

\vspace{2mm}
\noindent \textbf{Acknowledgements.}
{We thank David Fouhey for insightful discussions, and Saurabh Gupta and Tinghui Zhou for helpful comments. This work was supported in part by Intel/NSF VEC award IIS-1539099 and NSF Award IIS-1212798. We gratefully acknowledge NVIDIA corporation for the donation of GPUs used for this research.}

{\small
\bibliographystyle{ieee}
\bibliography{cvpr18dpr}
}

\clearpage
\setcounter{page}{1}
\renewcommand{\thepage}{\roman{page}}
\title{Appendix : Multi-view Consistency as Supervisory Signal \\ for Learning Shape and Pose Prediction
\vspace{-0.5cm}}
\author{Shubham Tulsiani, Alexei A. Efros, Jitendra Malik  \\
University of California, Berkeley\\
{\tt\small\{shubhtuls,efros,malik\}@eecs.berkeley.edu}
}
\setlength{\droptitle}{-2cm}
\maketitle
\vspace{-2mm}
\section*{A1. Loss Formulation}
\vspace{-2mm}
We briefly described, in the main text, the formulation of a view consistency loss $L(\bar{x},C; V)$ that measures the inconsistency between a shape $\bar{x}$ viewed according to camera $C$ and a depth/mask image $V$. Crucially, this loss was differentiable w.r.t both, pose and shape.
As indicated in the main text, our formulation builds upon previously proposed differentiable ray consistency formulation~\cite{drcTulsiani17} with some innovations to make it differentiable w.r.t pose. For presentation clarity, we first present our full formulation, and later discuss its relation to the previous techniques (a similar discussion can also be found in the main text).

\vspace{1mm}
\noindent \textbf{Notation.}
The (predicted) shape representation $\bar{x}$ is parametrized as occupancy probabilities of cells in a 3D grid. We use the convention that a particular value in the tensor $x$ corresponds to the probability of the corresponding voxel being $empty$. The verification image $V$ that we consider can be a depth or foreground mask image. Finally, the camera $C$ is parametrized via the intrinsic matrix $K$, and extrinsic matrix defined using a translation $t$ and rotation $R$.

\vspace{1mm}
\noindent \textbf{Per-pixel Error as Ray Consistency Cost.}
We consider the verification image $V$ one pixel at a time and define the per-pixel error using a (differentiable) ray consistency cost. Each pixel $p \equiv (u,v)$  has an associated value $v_p$ \eg in the case of a depth image, $v_p$ is the recorded depth at the pixel $p$. Additionally, each pixel corresponds to a ray originating from the camera centre and crossing the image plane at $(u,v)$. Given the camera parameters $C$ and shape $\bar{x}$, we can examine the ray corresponding to this pixel and check whether it is consistent with the observation $o_p$. We define a ray consistency cost function $L_p(\bar{x},C; v_p)$ to capture the error associated with the pixel $p$. The view consistency loss can then be defined as the sum of per-pixel errors $L(\bar{x}, C; V) \equiv \underset{p}{\sum} L_p(\bar{x}, C; v_p)$.


\vspace{1mm}
\noindent \textbf{Sampling Occupancies along a Ray.}
To define the consistency cost function $L_p(\bar{x}, C; v_p)$, we need to consider the ray as it is passing through the probabilistically occupied voxel grid $\bar{x}$. We do so by looking at discrete points sampled along the ray. Concretely, we sample points at a pre-defined set of $N = 80$ depth values $\{d_i | 1 \leq i \leq N\}$ along each ray. We denote by $x^p_i$ the occupancy value at the $i^{th}$ sample along this ray. To determine $x^p_i$, we look at the 3D coordinate of the corresponding point. Note that this can be determined using the camera parameters. Given the camera intrinsic parameters ($f_u,f_v,u_0,v_0$), the ray corresponding to the image pixel $(u,v)$ travels along the direction $(\frac{u-u_0}{f_u}, \frac{v-v_0}{f_v},1)$ in the camera frame. Therefore, the $i^{th}$ point along the ray, in the camera coordinate frame, is located at $l_i \equiv (\frac{u-u_0}{f_u}d_i, \frac{v-v_0}{f_v}d_i,d_i)$. Then, given the camera extrinsics $(R, t)$, we can compute the location of his point in the coordinate frame of the predicted shape $\bar{x}$. Finally, we can use trilinear sampling to determine the occupancy at this point by sampling the value at this using the occupancies $\bar{x}$. Denoting by $T(G,pt)$ a function that samples a volumetric grid $G$ at a location $pt$, we can compute the occupancy sampled at the $i^{th}$ as below.
\begin{gather}
x^p_i = \mathcal{T}( \bar{x} , R \times (l_i + t)~); \\
l_i \equiv (\frac{u-u_0}{f_u}d_i, \frac{v-v_0}{f_v}d_i,d_i)
\end{gather}
Note that since the trilinear sampling function $T$ is differentiable w.r.t its arguments, the sampled occupancy $x^p_i$ is differentiable w.r.t the shape $\bar{x}$ and the camera $C$.

\vspace{1mm}
\noindent \textbf{Probabilistic Ray Tracing.}
We have so far considered the ray associated with a pixel $p$ and computed samples with corresponding occupancy probabilities along it. We now trace this ray as it travels forward and use the samples along the ray as checkpoints. In particular, we assume that when the ray reaches the point corresponding to the $i^{th}$ sample, it either travels forward or terminates at that point. Conditioned on the ray reaching this sample, it travels forward with probability $x^p_i$ and terminates with likelihood $(1-x^p_i)$. We denote by $\mathbf{z}^p  \in \{1, \cdots , N+1\}$ a random variable corresponding to the sample index where the ray (probabilistically) terminates, where $z^p = N+1$ implies that the ray escapes. We call these probabilistic ray terminations as \emph{ray termination events} and can compute the probability distribution $q(z_p)$ for these.
\begin{gather}
\eqlabel{pzuv}
q(z^p = i) = (1 - x^p_i) \prod_{j=1}^{i-1} x^p_j~~~~\forall (i \leq N) ; \\ q(z^p = N+1) = \prod_{j=1}^{N} x^p_j;
\end{gather}

\noindent \textbf{Event Costs.} Each event corresponds to the ray terminating at a particular point. It is possible to assign a cost to each event based on how inconsistent it is to w.r.t the pixel value $v_p$. If we have a depth observation $v_p \equiv d_p$, we can penalize the event $z^p = i$ by measuring the difference between $d_p$ and $d_i$. Alternatively, if we have a foreground image observation \ie $v_p \equiv s_p \in \{0,1\}$ where $s_p = 1$ implies a foreground pixel, we can penalize all events which correspond to a different observation. We can therefore define a cost function $\psi_p(i)$ which computes the cost associated with event $z_p=i$.
\begin{gather}
\psi_p^{depth}(i) = |d_p - d_i|; \\
\psi_p^{mask}(i) = |s_p - \mathds{1}(i \leq N)|;
\end{gather}

\vspace{1mm}
\noindent \textbf{Ray Consistency Cost.}
We formulated the concept of ray termination events, and associated a probability and a cost to these. The ray consistency cost is then defined as the expected event cost.
\begin{gather}
L_p(\bar{x},C; v_p)~~=~~\underset{z_p}{\mathbb{E}}~\psi_p(z_p)~~=~~\sum_{i=1}^{N} q(z_p=i) \psi_p(i) 
\end{gather}
Note that the probabilities $q(z_p=i)$ are a differentiable function of $x_p$ which, in turn, is a differentiable function of shape $\bar{x}$ and camera $C$. The view consistency loss, which is simply a sum of multiple ray consistency terms, is therefore also differentiable w.r.t the shape and pose.

\vspace{1mm}
\textbf{Relation to Previous Work.} The formulation presented draws upon previous work on differentiable ray consistency~\cite{drcTulsiani17} and  leverages the notions of probabilistic ray termination events and event costs to define the ray consistency loss. A crucial difference however, is that we, using trilinear sampling, compute occupancies for point samples along the ray instead of directly using the occupancies of the voxels in the ray's path. Unlike their formulation, this allows our loss to also be differentiable w.r.t pose which is a crucial requirement for our scenario. Yan \etal~\cite{yan2016perspective} also use a similar sampling trick but their formulation is restricted to specifically using mask verification images and is additionally not leveraged for learning about pose. Tulsiani \etal~\cite{drcTulsiani17} also discuss how their formulation can be adapted to use more general verification images \eg color, semantics \etc using additional per-voxel predictions. While our experiments presented in the main text focus on leveraging mask or depth verification images, a similar generalization is possible for our formulation.

\section*{A2. Online Product Images Dataset}
\vspace{-2mm}
We used the `chair' object category from the Stanford Online Products Dataset~\cite{song16product}. To obtain associated foreground masks for these images, the semantic segmentation system from Chen \etal~\cite{chen2017deeplab}, where for each image, the mask was indicated by the pixels with most likely class label as `chair'. As the obtained segmentation masks were often incorrect, or objects in the images truncated/occluded, we manually selected images of unoccluded/untruncated instances with a reasonably accurate (though still noisy) predicted segmentation. For our training, we only used the object instances with atleast 2 valid views. This resulting dataset is visualized in \figref{ebay_vis}.
The result visualizations shown in the main text are using images from the original online products dataset~\cite{song16product}, but correspond to objects instances that were not used for our training (due to lack of a sufficient number of valid views).

\begin{figure*}[t!]
\centering
\includegraphics[width=0.98\textwidth]{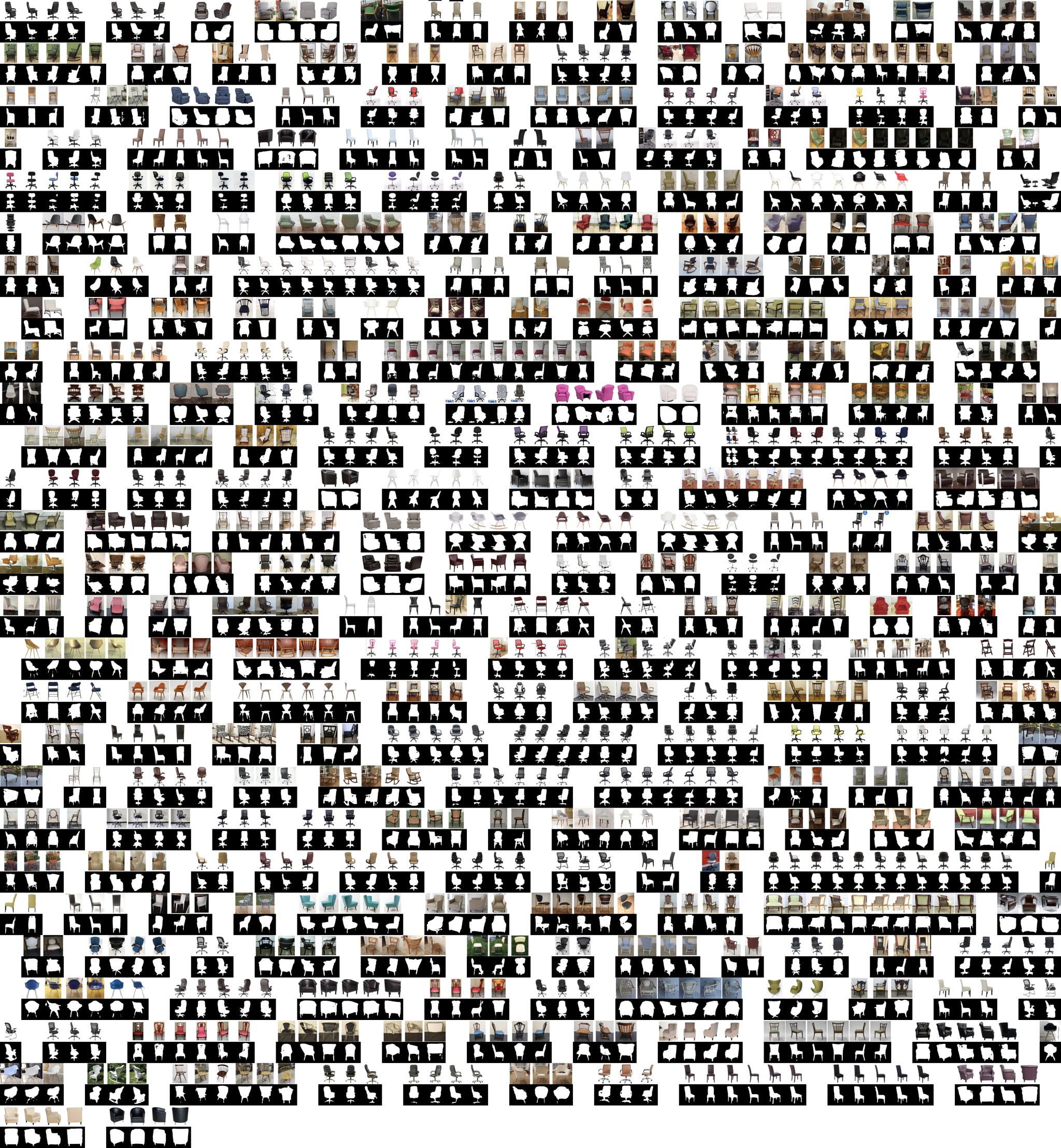}
\vspace{-2mm}
\caption{\small Training instances for the online products dataset. We visualize all the training images used along with their (approximate) segmentation masks, with images from the same object grouped together.}
\vspace{-2mm}
\figlabel{ebay_vis}
\end{figure*}

\end{document}